\documentclass[11pt]{article}

\usepackage[utf8]{inputenc}
\usepackage[T1]{fontenc}
\usepackage{lmodern}

\usepackage{amsmath, amssymb, amsthm}

\usepackage{graphicx}
\usepackage{booktabs}
\usepackage{multirow}
\usepackage{array}
\usepackage{adjustbox}
\usepackage{tabularx}
\usepackage{ragged2e}

\usepackage{geometry}
\usepackage{xurl}
\usepackage[hidelinks]{hyperref}

\hypersetup{
  pdftitle={Electric Vehicle Charging Load Forecasting: An Experimental Comparison of Machine Learning Methods},
  pdfauthor={Iason Kyriakopoulos; Yannis Theodoridis}
}

\title{Electric Vehicle Charging Load Forecasting: An Experimental Comparison of Machine Learning Methods}
\author{
Iason Kyriakopoulos\\
University of Piraeus, Greece\\
\texttt{iasonkyriakop@gmail.com}
\and
Yannis Theodoridis\\
University of Piraeus, Greece\\
\texttt{ytheod@unipi.gr}
}
\date{\today}

\begin{document}
\maketitle

\begin{abstract}
With the growing popularity of electric vehicles as a means of addressing climate change, concerns have emerged regarding their impact on electric grid management. As a result, predicting EV charging demand has become a timely and important research problem. This work investigates five forecasting models, from the statistical (ARIMA), machine learning (XGBoost), and deep learning families (GRU, LSTM, and Transformer), across short-, mid-, and long-term horizons (order of minutes, hours, and , days respectively) and three spatial scales (station-, region-, and city-level). Using datasets from Palo Alto, Boulder, Dundee, and Perth, we provide a systematic comparison of these methods. Results show that performance varies significantly across models and scales. For short-term regional and city forecasts, Transformers often achieve mean absolute errors (MAE) 10--15\% lower than classical baselines like ARIMA or XGBoost. On the other hand, when experimenting with mid- and long-term settings, recurrent neural networks (GRU, LSTM) consistently obtain the lowest errors; in some mid-term city-level cases, they reduce MAE by over 50\% compared to ARIMA. Furthermore,  XGBoost is competitive in localized scenarios, while ARIMA fails to scale to longer horizons or coarser aggregations. This study provides a reproducible benchmark for EV charging load forecasting across multiple scales. Our findings clarify trade-offs between statistical, machine learning, and deep learning methods, offering guidance on model selection for operational control versus long-term infrastructure planning.
\end{abstract}

\section{Introduction}
Electric Vehicles (EVs) have emerged as a key tool for mitigating climate change by reducing greenhouse gas emissions and dependence on fossil fuels, thereby supporting the transition toward more sustainable transportation systems~\cite{b1}. Their global adoption is rapidly accelerating, driven by international policies such as the Paris Agreement~\cite{b2}, which sets legally binding carbon neutrality targets and promotes the electrification of transport. As the number of EVs grows, so does electricity demand, creating challenges for grid management, energy planning, and infrastructure development. Forecasting EV charging consumption from individual stations up to city-wide levels is essential for ensuring a reliable energy supply~\cite{b3}, preventing overloads, and enabling the integration of renewable sources. Accurate forecasts also support grid stability by anticipating load fluctuations, optimizing the allocation of charging infrastructure, and informing energy and urban planning policies, making them fundamental for the development of smarter and cleaner cities~\cite{b4}.

Despite its importance, forecasting EV charging consumption remains a complex and demanding task \cite{b5}. The challenges stem from both data limitations and the inherent irregularity of EV usage patterns, as well as from methodological and evaluation issues. A central obstacle is the scarcity of open-source EV charging datasets \cite{b6}, with existing data varying widely in structure and scope, thereby complicating benchmarking and model comparison. Charging events also occur irregularly, particularly at the station level, making traditional time series forecasting methods less effective. External drivers, such as electricity prices, weather conditions, traffic, and other contextual factors, strongly influence charging demand \cite{b7}, yet are difficult to incorporate comprehensively into forecasting models. Consumption also displays layered temporal patterns and seasonality \cite{b8}, including daily (e.g., peak hours, daytime vs. night), weekly (e.g., weekday vs. weekend), and yearly (e.g., winter vs. summer) cycles, which require models capable of capturing such overlapping dynamics. Moreover, the rapid growth in EV adoption continuously reshapes usage patterns, necessitating forecasting approaches that can adapt to evolving trends. Finally, predictions are required at different levels of aggregation, from single stations to entire cities, each presenting distinct patterns and challenges that call for flexible and scalable modeling frameworks.

The main objective of this research is therefore to conduct a systematic comparison of several widely used forecasting models applied to EV charging consumption data. The study evaluates how these models perform across different temporal resolutions, with the goal of drawing meaningful conclusions about their applicability, strengths, and limitations in this domain.

To this end, we forecast EV charging consumption using five models that span classical statistical approaches (ARIMA), machine learning (XGBoost), and deep learning (GRU, LSTM, Transformer). These models are trained on four popular real-world EV charging datasets from Palo Alto, Boulder, Dundee, and Perth, and evaluated at multiple aggregation levels, from individual charging stations to regional and city-wide consumption. The comparison covers three forecasting regimes: short term (up to 30 minutes ahead, in 10-minute steps), mid term (up to 8 hours ahead, in 2-hour steps), and long term (up to 5 days ahead, in 1-day steps).

The main contributions of this work are summarized as follows:
\begin{itemize}
    \item \textbf{Multi-scale EV charging benchmark.} We design a unified experimental framework that evaluates five widely used forecasting models (ARIMA, XGBoost, GRU, LSTM, Transformer) across three temporal resolutions (10-minute, hourly, and daily) and three spatial aggregation levels (station, region, city) in four real-world EV charging datasets.
    \item \textbf{Consistent and reproducible evaluation protocol.} All models are trained and evaluated under a common preprocessing, feature engineering, and recursive multi-step forecasting protocol. The full codebase and links to the datasets are publicly available, facilitating reproducibility and fair comparison. \footnote{For reproducibility purposes, the entire experimental protocol (data preprocessing, model training, inference, and evaluation pipelines), together with links to datasets, is available at \url{https://github.com/DataStories-UniPi/electric-vehicle-charging-load-forecasting}.}
    \item \textbf{Systematic comparison across horizons and scales.} We provide a detailed comparison of forecasting performance across short-, mid-, and long-term horizons, highlighting how model rankings change with both prediction horizon and spatial aggregation level.
    \item \textbf{Insights on model families.} We show that Transformer models tend to dominate short-term forecasting at regional and city aggregation levels, whereas recurrent neural networks, particularly GRU and LSTM, consistently provide the most accurate mid- and long-term forecasts in most datasets. We also document cases where Transformers degrade markedly (e.g., Dundee, long-term horizons) and discuss possible causes.
\end{itemize}

The remainder of the paper is organized as follows. Section~\ref{sec:problem} formalizes the EV charging load forecasting problem and outlines the forecasting methods considered. Section~\ref{sec:framework} presents the datasets, the preprocessing pipeline, and the performance evaluation framework. Section~\ref{sec:experiments} describes the experimental setup and reports the results with comparative analysis and discussion. Section~\ref{sec:related-work} reviews directly comparable related work. Section~\ref{sec:conclusion} concludes and outlines directions for future research.

\section{The EV Charging Load Forecasting Problem and Related Methods}
\label{sec:problem}
\subsection{Problem Definition in Terms of Time Series Forecasting}

The primary task of this study is to forecast the total energy consumption (in kWh) of EV charging infrastructure over future time intervals. Formally, the problem at hand is formulated as follows:

\textbf{The EV Charging Load Forecasting Problem:} Given a historical sequence of aggregated charging energy consumption values \[ X = \{x_1, x_2, \ldots, x_T\}, \] where \(x_t\) represents the total energy consumed at time step \(t\), the objective is to train a model that can predict future consumption values \[ \hat{X} = \{\hat{x}_{T+1}, \hat{x}_{T+2}, \ldots, \hat{x}_{T+H}\}, \] over a forecasting horizon \(H\). Depending on the application, \(H\) could correspond to short- (order of minutes), mid- (order of hours), or long-term (order of days) predictions. These horizons enable the evaluation of forecasting models under varying temporal complexities, ranging from near real-time operational to strategic planning decisions.

\begin{figure*}[!t]
\centering
\includegraphics[width=0.85\textwidth]{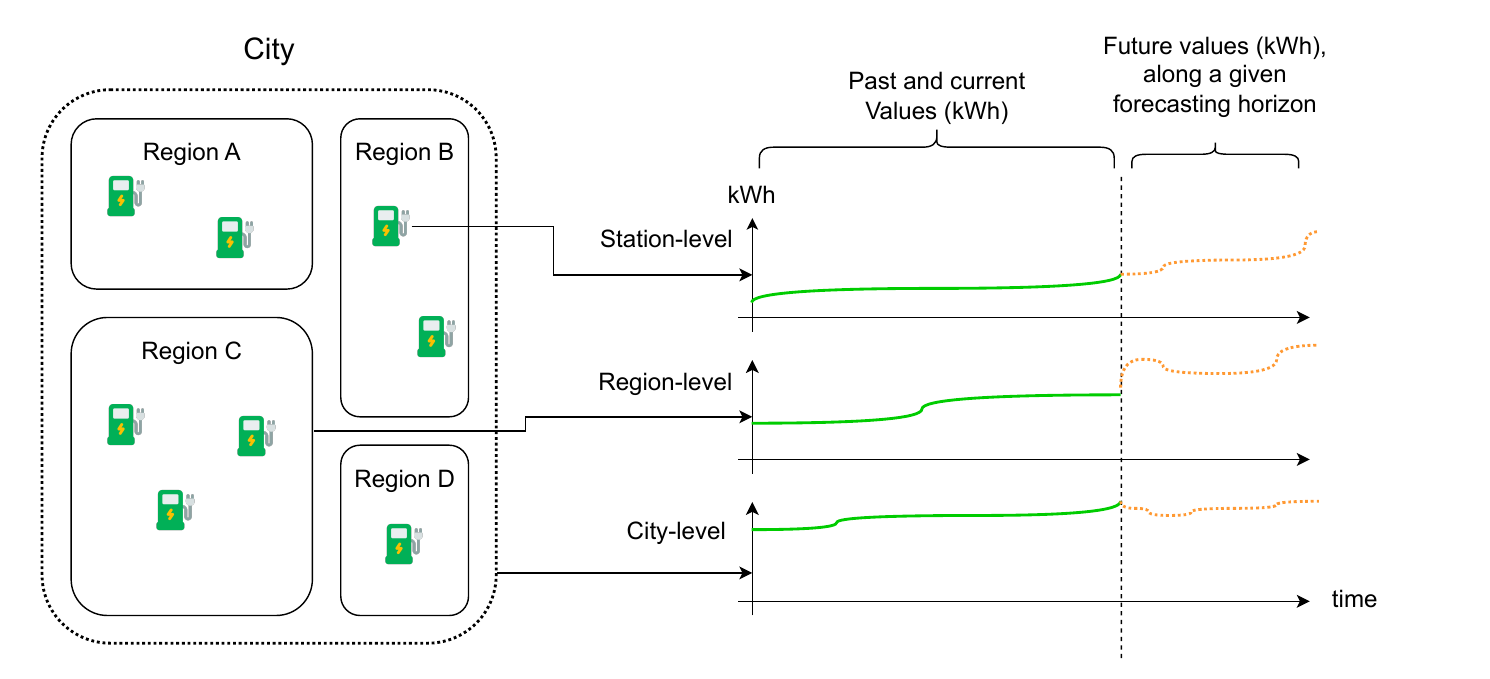}
\caption{Illustration of the EV charging load forecasting problem.}
\label{fig:EVproblem}
\end{figure*}

As implied by the problem definition above, forecasting can be conducted at multiple spatial aggregation levels. At the charging station level, each individual station is treated as a separate time series.At the regional level, groups of stations within a given area are combined, and their total energy consumption is aggregated to reflect regional demand. At the city level, all stations within a city are aggregated into a single time series representing overall city-wide load. This multi-level structure is illustrated in Fig.~\ref{fig:EVproblem}.

By combining multiple prediction horizons with different spatial resolutions, this study establishes a multi-scale forecasting framework. This framework enables systematic evaluation of model performance across fine-grained local scales and broader aggregated levels, while also accounting for short-term fluctuations and longer-term trends.

\subsection{Forecasting Methods Considered}
This study compares forecasting models spanning traditional statistical approaches, machine learning, and deep learning.
Rather than attempting to cover all available architectures in the literature, we focus on a set of widely used, well-understood baselines that are representative of major model families in time series forecasting.

ARIMA (Autoregressive Integrated Moving Average) serves as a baseline due to its simplicity, interpretability, and long-standing use in time series analysis \cite{b9}.
XGBoost (Extreme Gradient Boosting) is a tree-based ensemble method capable of efficiently modeling non-linear relationships, providing a strong benchmark against both statistical and deep learning approaches \cite{b10}.
Gated Recurrent Units (GRU) are recurrent neural networks designed to handle sequential data through gating mechanisms, mitigating vanishing gradient issues and capturing medium- to long-term dependencies \cite{b11}.
Long Short-Term Memory (LSTM) networks extend traditional RNNs with memory cells and gating mechanisms, enabling the modeling of long-range temporal dependencies, and remain widely used in time series forecasting as robust deep learning architectures \cite{b12}.
Based on self-attention mechanisms, Transformers can capture dependencies across entire sequences in parallel and are increasingly applied in time series forecasting \cite{b13}.

These five models were chosen because (i) they are considered typical reference points in both the EV charging and broader load-forecasting literature, (ii) they cover complementary modeling paradigms (linear statistical, tree-based ensemble, and sequential deep learning), and (iii) they are readily available in mature open-source implementations, which facilitates reproducibility.

Prior research highlights the strengths and limitations of these forecasting approaches, but notable gaps remain. Few studies systematically evaluate how models trained at the station level scale to regional or city-level consumption, leaving open questions about multi-scale adaptability. Moreover, many works assess performance at a single forecast horizon, and standardized comparisons across short-, mid-, and long-term horizons are rare. This study primarily addresses these two challenges: (i) scalability across spatial levels (from individual stations to city-wide aggregation) and (ii) systematic performance comparison across multiple forecasting horizons.

\section{The Experimental Comparison Framework}
\label{sec:framework}
\begin{figure*}[!t]
\centering
\includegraphics[width=0.85\textwidth]{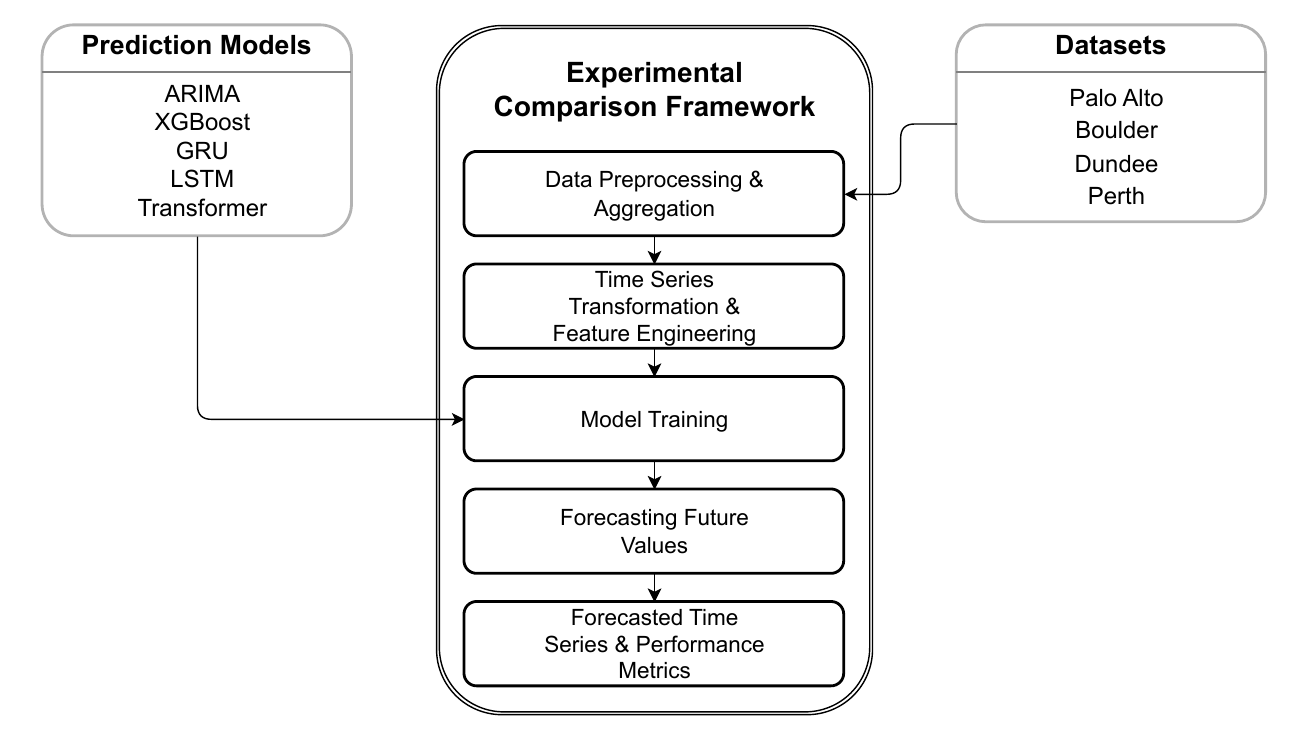}
\caption{Overview of the experimental comparison framework.}
\label{fig:comparison_framework}
\end{figure*}

This section outlines the experimental framework used to compare forecasting models under consistent conditions. It describes how real-world EV charging datasets are standardized and transformed into time series suitable for model training, and how model performance is evaluated across multiple spatial aggregation levels and temporal horizons. The framework ensures methodological uniformity, enabling meaningful comparison of statistical, machine learning, and deep learning approaches. A bird’s-eye view of the framework architecture is illustrated in Fig. \ref{fig:comparison_framework}.

\subsection{Datasets and Preprocessing}
This study utilizes four popular real-world datasets of EV charging sessions collected from different cities, namely, Palo Alto, Boulder, Dundee, and Perth \cite{b14}. Each dataset contains individual charging sessions, with varying duration, spatial coverage, granularity, and associated metadata. Table~\ref{tab:datasets} provides an overview of the dataset characteristics. To ensure consistency and comparability across these diverse datasets, we develop and apply a unified preprocessing pipeline, as described below.

\begin{table*}[!t]
\centering
\caption{Overview of EV Charging Datasets}
\label{tab:datasets}
\small
\renewcommand{\arraystretch}{1.25}
\setlength{\tabcolsep}{4pt}

\begin{tabularx}{\textwidth}{%
>{\RaggedRight\arraybackslash}X
>{\RaggedRight\arraybackslash}X
c c
>{\RaggedRight\arraybackslash}X}
\toprule
\textbf{City} &
\textbf{Time Range} &
\textbf{Nr. of Stations} &
\textbf{Nr. of Regions} &
\textbf{Nr. of Charging Session Records} \\
\midrule
Palo Alto, CA (USA) & July 2011 -- Dec.\ 2020 & 47 & 3  & 259,415 \\
Boulder, CO (USA)  & Jan.\ 2018 -- Mar.\ 2021 & 27 & 5  & 24,081 \\
Dundee (UK)        & Jan.\ 2017 -- Dec.\ 2018 & 67 & 34 & 52,752 \\
Perth (Australia)  & Jan.\ 2016 -- Dec.\ 2019 & 36 & 22 & 66,664 \\
\bottomrule
\end{tabularx}
\end{table*}

Each dataset originally contained a mix of core and auxiliary fields. The columns retained across all cities included the start and end times of each charging session (merged from separate date and time fields when necessary), energy consumption in kWh per session, a station identifier, and a regional grouping field (e.g., ZIP code or site label). Auxiliary fields that were either irrelevant for forecasting or inconsistently available, such as port type, greenhouse gas emissions saved, session ID, and user ID, were excluded.

A small fraction of missing values was present. To maintain simplicity and avoid introducing additional assumptions, rows with missing entries were removed, resulting in minimal data loss. Data integrity checks were also performed to exclude invalid sessions for the purpose of our study, such as those reporting negative energy consumption or end times preceding start times.

Time zones were standardized for consistency. Datasets with timezone-aware timestamps were converted to Coordinated Universal Time (UTC), while datasets lacking explicit timezone metadata were assumed to follow the local time of the respective city.

To transform session-based data into a time series format, energy consumption was aggregated into fixed 10-minute intervals per station by summing energy from all sessions overlapping each interval. This 10-minute resolution served as the base granularity, from which hourly and daily time series were derived through resampling.

Additional time-based features were engineered, including calendar variables such as holidays, weekends, day of the week, and month of the year. To capture temporal dependencies, multiple lagged versions of the target variable were generated at different offsets appropriate to each temporal resolution (10-minute, hourly, and daily). This ensured that both short-term fluctuations and longer-term seasonal patterns were represented in the feature space.

Finally, energy consumption values were standardized using z-score normalization. Forecasts were evaluated directly in the normalized domain, without inverse transformation, to enable consistent comparison across datasets and with prior studies (see Section~\ref{sec:related-work}).

\subsection{The Performance Evaluation Protocol}
To ensure a fair and consistent comparison, a unified evaluation protocol is applied. Each model is evaluated across all short-, mid-, and long-term forecasting horizons and across all spatial aggregation levels (station, region, and city). Evaluation is performed independently for each dataset, producing one set of performance scores per dataset for every combination of forecasting horizon and spatial aggregation level. Results are reported per dataset, with the best-performing model highlighted for each combination of dataset, spatial aggregation level, forecasting horizon, and error metric.

Performance is assessed using two standard error metrics, namely, MAE, which provides an intuitive measure of average error magnitude, and RMSE, which penalizes larger errors more strongly and is therefore sensitive to outliers. These two metrics were selected for their simplicity, interpretability, and widespread use in time series forecasting literature~\cite{b7}. Other metrics such as MAPE were avoided due to instability when actual values approach zero \cite{b15}, while $R^2$ was excluded as it is less interpretable in non-linear forecasting contexts \cite{b16}.

\section{Experimental Study}
\label{sec:experiments}

\subsection{Experimental Setup}
This section describes the technical environment and modeling configurations used in the experiments.

Two distinct modeling strategies were adopted in this study. The first strategy employed multi-station learning, where a single model was trained per city and temporal resolution using pooled data from all stations. This approach was applied to XGBoost, GRU, LSTM, and Transformer architectures. The input space combined one-hot encodings of station and region identifiers with calendar and holiday indicators. All lagged features and target values were standardized using z-score normalization computed from the training partition. Forecasting was conducted in a recursive walk-forward manner, where predictions at each step were iteratively fed back as inputs, thereby updating the lag state of each station over time.  This recursive walk-forward strategy is operationally realistic for EV charging applications, where forecasts are continually updated as new observations arrive. However, it also implies that errors made at earlier steps can propagate to later ones, especially for mid- and long-term horizons. In our setting, this effect is partly mitigated by temporal aggregation (from 10-minute to hourly and daily series) and by training all models under the same recursive protocol, so that error accumulation affects them in a comparable way.

The second strategy relied on per-station univariate ARIMA modeling. Here, each charging station was treated independently, and ARIMA models were fitted to standardized series of energy consumption. To constrain computational cost, order selection was performed using a restricted grid search over $(p,d,q)$ with $p,q \in \{0,1,2\}$ and $d \in \{0,1\}$, selecting the specification with the lowest AIC (Akaike Information Criterion). Training was limited to a capped recent window of the data: 2{,}048 points for 10-minute, 1{,}536 for hourly, and 730 for daily resolution (corresponding roughly to two weeks, two months, and two years of history, respectively), which balances computational feasibility with the amount of recent information typically sufficient for reliable ARIMA estimation. Stationarity and invertibility were enforced, and when estimation failed, a naïve persistence model was substituted.

All models were trained and evaluated over three temporal resolutions:
\begin{itemize}
    \item \textbf{Short-term forecasting}: 10 up to 30 minutes, in steps of 10 minutes (i.e., 3 steps);
    \item \textbf{Mid-term forecasting}: 2 up to 8 hours, in steps of 2 hours (i.e., 4 steps); and
    \item \textbf{Long-term forecasting}: 1 up to 5 days, in steps of 1 day (i.e., 5 steps).
\end{itemize}

Hyperparameters for the learning models were fixed rather than exhaustively tuned, in order to ensure reproducibility and maintain computational feasibility across datasets. GPU acceleration was used for XGBoost and neural models, while ARIMA was executed on CPU. The settings were as follows:
\begin{itemize}
  \item \textbf{ARIMA}: per-station order selected from the restricted AIC grid as described above, trained on the capped tail of the training set, and used to forecast the full test horizon.
  \item \textbf{XGBoost}: {\texttt{\fontsize{9}{10}\selectfont device=cuda, tree\_method=hist}}; learning rate 0.05; maximum depth 8; subsample 0.8; {\texttt{\fontsize{9}{10}\selectfont colsample\_bytree}} 0.8; $\ell_2$ regularization 1.0; up to 2000 rounds with early stopping (patience 200).
  \item \textbf{GRU/LSTM}: sequence length equal to the number of engineered lag features; 64 hidden units; Adam optimizer ($10^{-3}$); batch size 2048; maximum 200 epochs with early stopping (patience 20); TensorFlow mixed precision enabled on GPU.
  \item \textbf{Transformer}: $d_{\text{model}}{=}128$, 8 attention heads, 4 layers, feed-forward dimension 256, dropout 0.1; trained with the same optimizer, batch size, early stopping, and mixed precision settings, as with GRU/LSTM.
\end{itemize}
The hyperparameters above were chosen to be conservative and stable across datasets, inspired by settings commonly reported in the time series forecasting literature and by preliminary convergence checks, rather than by dataset-specific tuning or automated search. This trades off some potential performance gains in favor of comparability and computational feasibility across the large number of experimental configurations (cities, aggregation levels, horizons). Consequently, the reported scores should be interpreted as conservative estimates of what each model family can achieve under a simple, uniform configuration rather than as fully optimized best-case results.

All experiments were conducted in Python, with version 3.10 used for data preprocessing 
and version 3.8 for forecasting, model training and inference. The models were developed 
using standard open-source libraries, including 
\texttt{\fontsize{9}{10}\selectfont pandas}, 
\texttt{\fontsize{9}{10}\selectfont numpy}, 
\texttt{\fontsize{9}{10}\selectfont statsmodels}, 
\texttt{\fontsize{9}{10}\selectfont scikit-learn}, 
and \texttt{\fontsize{9}{10}\selectfont tensorflow}.

Model training and evaluation were performed on a single NVIDIA A100 GPU node with 1~TB of system memory. Leveraging GPU acceleration significantly reduced training times for deep learning models, enabling efficient experimentation across all datasets.

\subsection{Experimental Results}
In the following paragraphs, we present the performance of each of the five models across the different forecasting horizons and spatial aggregation levels.

Tables~2--4 summarize model performance for short-, mid-, and long-term horizons at different spatial aggregation levels and for each dataset.
In particular, Table~\ref{tab:shortterm} presents short-term (10-30 minutes) results, Table~\ref{tab:midterm} shows mid-term (2-8 hours) outcomes, and Table~\ref{tab:longterm} reports long-term (1-5 days) forecasting accuracy. 
Each table reports detailed MAE and RMSE values per model, dataset, aggregation level, and horizon, while the accompanying text focuses on relative rankings and dominant patterns rather than individual scores. In practice, practitioners are often more interested in which model families tend to perform best in a given regime (e.g., short-term station-level versus long-term city-level) than in small numerical differences. For this reason, we emphasize model rankings and consistent trends across horizons and datasets, and refer to the tables for exact values.

\begin{table}[htbp]
\centering
\caption{Forecasting results for up to 30-minute horizon (10-minute steps). Datasets: Palo Alto, Boulder, Dundee, Perth. Lower is better.}
\label{tab:shortterm}
\scriptsize
\setlength{\tabcolsep}{4pt}
\renewcommand{\arraystretch}{1.05}

\begin{tabular}{l l l c c c c c c c c c}
\toprule
\textbf{Model} & \textbf{Dataset} & \textbf{Metric} &
\multicolumn{3}{c}{\textbf{10 minutes}} &
\multicolumn{3}{c}{\textbf{20 minutes}} &
\multicolumn{3}{c}{\textbf{30 minutes}} \\
\cmidrule(lr){4-6}\cmidrule(lr){7-9}\cmidrule(lr){10-12}
& & &
\textbf{Station} & \textbf{Region} & \textbf{City} &
\textbf{Station} & \textbf{Region} & \textbf{City} &
\textbf{Station} & \textbf{Region} & \textbf{City} \\
\midrule

\multirow{8}{*}{ARIMA}
& \multirow{2}{*}{Palo Alto} & MAE  & 0.92 & 6.34 & 13.12 & 0.91 & 6.33 & 13.09 & 0.91 & 6.31 & 13.06 \\
&                           & RMSE & 1.19 & 8.14 & 15.54 & 1.18 & 8.11 & 15.49 & 1.17 & 8.09 & 15.45 \\
& \multirow{2}{*}{Boulder}  & MAE  & 0.26 & 0.85 & 3.11  & 0.26 & 0.84 & 3.06  & 0.26 & 0.83 & 3.04  \\
&                           & RMSE & 0.68 & \textbf{1.50} & 4.14  & \textbf{0.66} & \textbf{1.47} & 4.08  & \textbf{0.65} & \textbf{1.45} & 4.04  \\
& \multirow{2}{*}{Dundee}   & MAE  & \textbf{0.62} & 1.26 & 18.16 & \textbf{0.61} & 1.25 & 17.58 & \textbf{0.61} & 1.25 & 17.30 \\
&                           & RMSE & 1.43 & 2.19 & 23.81 & 1.28 & 2.05 & 22.66 & 1.20 & 1.97 & 21.99 \\
& \multirow{2}{*}{Perth}    & MAE  & 0.62 & 1.60 & 12.02 & 0.61 & 1.59 & 11.62 & 0.61 & 1.58 & 11.37 \\
&                           & RMSE & 1.36 & 2.91 & 13.78 & 1.26 & 2.72 & 13.18 & 1.20 & 2.61 & 12.83 \\
\midrule

\multirow{8}{*}{XGBoost}
& \multirow{2}{*}{Palo Alto} & MAE  & \textbf{0.88} & 5.67 & 14.71 & \textbf{0.87} & 5.64 & 14.67 & \textbf{0.87} & 5.62 & 14.64 \\
&                           & RMSE & \textbf{1.02} & 6.38 & 16.28 & \textbf{1.01} & 6.35 & 16.24 & \textbf{1.00} & 6.32 & 16.20 \\
& \multirow{2}{*}{Boulder}  & MAE  & 0.87 & 3.29 & 14.80 & 0.86 & 3.28 & 14.80 & 0.86 & 3.26 & 14.79 \\
&                           & RMSE & 0.98 & 3.47 & 15.52 & 0.96 & 3.45 & 15.50 & 0.96 & 3.44 & 15.49 \\
& \multirow{2}{*}{Dundee}   & MAE  & 1.37 & 2.69 & 55.99 & 1.31 & 2.61 & 55.80 & 1.26 & 2.57 & 55.70 \\
&                           & RMSE & 1.67 & 2.97 & 58.80 & 1.56 & 2.88 & 58.40 & 1.49 & 2.82 & 58.16 \\
& \multirow{2}{*}{Perth}    & MAE  & 0.94 & 2.27 & 21.52 & 0.91 & 2.20 & 21.12 & 0.90 & 2.14 & 20.88 \\
&                           & RMSE & 1.36 & 2.81 & 23.33 & 1.26 & 2.64 & 23.00 & 1.20 & 2.54 & 22.81 \\
\midrule

\multirow{8}{*}{GRU}
& \multirow{2}{*}{Palo Alto} & MAE  & 1.05 & 6.07 & 16.55 & 1.04 & 6.04 & 16.51 & 1.04 & 6.02 & 16.48 \\
&                           & RMSE & 1.15 & 6.72 & 18.14 & 1.13 & 6.69 & 18.10 & 1.13 & 6.67 & 18.07 \\
& \multirow{2}{*}{Boulder}  & MAE  & 0.32 & 0.97 & 2.81  & 0.31 & 0.95 & 2.76  & 0.31 & 0.94 & 2.72  \\
&                           & RMSE & 0.68 & 1.59 & \textbf{4.09}  & 0.67 & 1.55 & \textbf{4.02}  & \textbf{0.65} & 1.53 & \textbf{3.97}  \\
& \multirow{2}{*}{Dundee}   & MAE  & 1.08 & 1.89 & 33.22 & 1.03 & 1.81 & 32.75 & 0.99 & 1.76 & 32.48 \\
&                           & RMSE & 1.45 & 2.25 & 35.60 & 1.33 & 2.13 & 34.95 & 1.25 & 2.06 & 34.56 \\
& \multirow{2}{*}{Perth}    & MAE  & 0.85 & 2.03 & 17.67 & 0.83 & 1.95 & 17.39 & 0.81 & 1.90 & 17.24 \\
&                           & RMSE & 1.43 & 2.88 & 20.54 & 1.33 & 2.71 & 20.17 & 1.27 & 2.61 & 19.95 \\
\midrule

\multirow{8}{*}{LSTM}
& \multirow{2}{*}{Palo Alto} & MAE  & 1.00 & 5.87 & 15.87 & 0.99 & 5.85 & 15.84 & 0.98 & 5.83 & 15.81 \\
&                           & RMSE & 1.10 & 6.53 & 17.46 & 1.09 & 6.50 & 17.42 & 1.08 & 6.47 & 17.38 \\
& \multirow{2}{*}{Boulder}  & MAE  & 0.54 & 1.90 & 7.11  & 0.53 & 1.89 & 7.07  & 0.53 & 1.88 & 7.04  \\
&                           & RMSE & 0.91 & 2.57 & 8.70  & 0.90 & 2.55 & 8.67  & 0.89 & 2.53 & 8.64  \\
& \multirow{2}{*}{Dundee}   & MAE  & 1.49 & 2.89 & 64.93 & 1.44 & 2.81 & 64.78 & 1.40 & 2.75 & 64.69 \\
&                           & RMSE & 1.85 & 3.20 & 68.71 & 1.72 & 3.09 & 68.36 & 1.65 & 3.02 & 68.14 \\
& \multirow{2}{*}{Perth}    & MAE  & \textbf{0.46} & \textbf{1.07} & \textbf{7.97}  & \textbf{0.44} & \textbf{1.03} & \textbf{7.54}  & \textbf{0.43} & \textbf{0.99} & \textbf{7.27}  \\
&                           & RMSE & \textbf{1.06} & \textbf{2.09} & \textbf{10.98} & \textbf{0.95} & \textbf{1.88} & \textbf{10.27} & \textbf{0.88} & \textbf{1.74} & \textbf{9.84}  \\
\midrule

\multirow{8}{*}{Transformer}
& \multirow{2}{*}{Palo Alto} & MAE  & 0.93 & \textbf{5.17} & \textbf{12.84} & 0.92 & \textbf{5.15} & \textbf{12.80} & 0.91 & \textbf{5.13} & \textbf{12.77} \\
&                           & RMSE & 1.11 & \textbf{5.98} & \textbf{14.34} & 1.10 & \textbf{5.94} & \textbf{14.29} & 1.09 & \textbf{5.91} & \textbf{14.25} \\
& \multirow{2}{*}{Boulder}  & MAE  & \textbf{0.22} & \textbf{0.76} & \textbf{2.71}  & \textbf{0.21} & \textbf{0.76} & \textbf{2.67}  & \textbf{0.21} & \textbf{0.75} & \textbf{2.65}  \\
&                           & RMSE & \textbf{0.67} & 1.58 & 4.34  & \textbf{0.66} & 1.54 & 4.27  & \textbf{0.65} & 1.52 & 4.22  \\
& \multirow{2}{*}{Dundee}   & MAE  & 0.73 & \textbf{1.17} & \textbf{14.52} & 0.68 & \textbf{1.09} & \textbf{13.70} & 0.63 & \textbf{1.04} & \textbf{13.22} \\
&                           & RMSE & \textbf{1.18} & \textbf{1.70} & \textbf{18.54} & \textbf{1.04} & \textbf{1.56} & \textbf{17.22} & \textbf{0.95} & \textbf{1.46} & \textbf{16.43} \\
& \multirow{2}{*}{Perth}    & MAE  & 1.53 & 3.79 & 42.94 & 1.49 & 3.70 & 42.91 & 1.46 & 3.65 & 42.90 \\
&                           & RMSE & 1.89 & 4.30 & 45.40 & 1.81 & 4.18 & 45.23 & 1.77 & 4.10 & 45.13 \\
\bottomrule

\end{tabular}
\end{table}

\begin{table}[htbp]
\centering
\caption{Forecasting results for up to 8-hour horizon (2-hour steps). Datasets: Palo Alto, Boulder, Dundee, Perth. Lower is better.}
\label{tab:midterm}
\scriptsize
\setlength{\tabcolsep}{3.5pt}
\renewcommand{\arraystretch}{1.05}

\begin{tabular}{l l l c c c c c c c c c c c c}
\toprule
\textbf{Model} & \textbf{Dataset} & \textbf{Metric} &
\multicolumn{3}{c}{\textbf{2 hours}} &
\multicolumn{3}{c}{\textbf{4 hours}} &
\multicolumn{3}{c}{\textbf{6 hours}} &
\multicolumn{3}{c}{\textbf{8 hours}} \\
\cmidrule(lr){4-6}\cmidrule(lr){7-9}\cmidrule(lr){10-12}\cmidrule(lr){13-15}
& & &
\textbf{St.} & \textbf{Reg.} & \textbf{City} &
\textbf{St.} & \textbf{Reg.} & \textbf{City} &
\textbf{St.} & \textbf{Reg.} & \textbf{City} &
\textbf{St.} & \textbf{Reg.} & \textbf{City} \\
\midrule

\multirow{8}{*}{ARIMA}
& \multirow{2}{*}{Palo Alto} & MAE  & 0.74 & 5.10 & 13.30 & 0.72 & 4.92 & 12.81 & 0.70 & 4.71 & 12.50 & 0.69 & 4.49 & 11.21 \\
&                           & RMSE & 1.00 & 6.90 & 17.35 & 0.96 & 6.66 & 16.90 & 0.91 & 6.47 & 16.57 & 0.87 & 6.14 & 15.55 \\
& \multirow{2}{*}{Boulder}  & MAE  & \textbf{0.28} & \textbf{0.85} & 3.08  & \textbf{0.27} & \textbf{0.79} & 2.83  & \textbf{0.27} & \textbf{0.75} & 2.68  & \textbf{0.26} & \textbf{0.72} & 2.42  \\
&                           & RMSE & \textbf{0.63} & \textbf{1.43} & 4.08  & \textbf{0.56} & \textbf{1.27} & 3.73  & \textbf{0.51} & \textbf{1.16} & 3.43  & 0.47 & \textbf{1.08} & 3.24  \\
& \multirow{2}{*}{Dundee}   & MAE  & 1.09 & 1.91 & 31.66 & 1.07 & 1.86 & 30.77 & 1.06 & 1.84 & 30.23 & 1.06 & 1.83 & 29.88 \\
&                           & RMSE & 1.53 & 2.42 & 36.00 & 1.43 & 2.32 & 35.28 & 1.38 & 2.25 & 34.88 & 1.34 & 2.21 & 34.60 \\
& \multirow{2}{*}{Perth}    & MAE  & 0.93 & 2.18 & 19.02 & 0.90 & 2.07 & 18.25 & 0.88 & 2.01 & 17.22 & 0.86 & 1.98 & 16.50 \\
&                           & RMSE & 1.35 & 2.83 & 21.34 & 1.22 & 2.59 & 20.68 & 1.14 & 2.44 & 19.82 & 1.10 & 2.36 & 19.22 \\
\midrule

\multirow{8}{*}{XGBoost}
& \multirow{2}{*}{Palo Alto} & MAE  & 0.59 & 3.22 & 7.81  & 0.45 & 2.98 & 7.40  & 0.42 & 2.84 & 7.16  & \textbf{0.39} & 2.67 & 6.82  \\
&                           & RMSE & \textbf{0.74} & 3.93 & 9.28  & \textbf{0.60} & 3.59 & 8.75  & \textbf{0.54} & 3.41 & 8.46  & \textbf{0.49} & 3.20 & 8.09  \\
& \multirow{2}{*}{Boulder}  & MAE  & 0.57 & 1.92 & 7.80  & 0.54 & 1.84 & 7.74  & 0.51 & 1.78 & 7.69  & 0.49 & 1.75 & 7.67  \\
&                           & RMSE & 0.75 & 2.22 & 8.43  & 0.68 & 2.09 & 8.27  & 0.63 & 1.99 & 8.12  & 0.58 & 1.94 & 8.08  \\
& \multirow{2}{*}{Dundee}   & MAE  & 1.16 & 2.40 & 53.64 & 1.08 & 2.30 & 53.44 & 1.05 & 2.26 & 53.51 & 1.02 & 2.23 & 53.46 \\
&                           & RMSE & 1.38 & 2.70 & 56.18 & 1.26 & 2.57 & 55.82 & 1.20 & 2.50 & 55.58 & 1.17 & 2.46 & 55.50 \\
& \multirow{2}{*}{Perth}    & MAE  & 0.63 & 1.42 & 11.73 & 0.56 & 1.26 & 10.96 & 0.51 & 1.16 & 10.16 & 0.49 & 1.10 & 10.09 \\
&                           & RMSE & 0.94 & 1.89 & 13.80 & 0.78 & 1.60 & 13.02 & 0.69 & 1.43 & 12.33 & 0.64 & 1.35 & 12.16 \\
\midrule

\multirow{8}{*}{GRU}
& \multirow{2}{*}{Palo Alto} & MAE  & 0.63 & 3.63 & 9.01  & 0.54 & 3.36 & 8.48  & 0.50 & 3.20 & 8.23  & 0.47 & 3.05 & 7.95  \\
&                           & RMSE & 0.79 & 4.76 & 11.99 & 0.70 & 4.43 & 11.45 & 0.63 & 4.15 & 10.84 & 0.58 & 3.85 & 10.13 \\
& \multirow{2}{*}{Boulder}  & MAE  & 0.70 & 2.52 & 11.03 & 0.67 & 2.43 & 10.97 & 0.65 & 2.37 & 10.94 & 0.62 & 2.33 & 10.93 \\
&                           & RMSE & 0.96 & 3.07 & 12.01 & 0.90 & 2.95 & 11.82 & 0.84 & 2.84 & 11.74 & 0.79 & 2.78 & 11.67 \\
& \multirow{2}{*}{Dundee}   & MAE  & \textbf{0.78} & \textbf{1.36} & \textbf{20.36} & \textbf{0.69} & \textbf{1.26} & \textbf{19.61} & \textbf{0.65} & \textbf{1.20} & \textbf{19.03} & \textbf{0.62} & \textbf{1.17} & \textbf{18.91} \\
&                           & RMSE & \textbf{1.03} & \textbf{1.75} & \textbf{23.23} & \textbf{0.90} & \textbf{1.60} & \textbf{22.21} & \textbf{0.83} & \textbf{1.50} & \textbf{21.46} & \textbf{0.79} & \textbf{1.46} & \textbf{21.30} \\
& \multirow{2}{*}{Perth}    & MAE  & \textbf{0.53} & \textbf{1.17} & \textbf{8.17}  & \textbf{0.47} & \textbf{1.03} & \textbf{7.43}  & \textbf{0.43} & \textbf{0.94} & \textbf{6.76}  & \textbf{0.41} & \textbf{0.89} & \textbf{6.54}  \\
&                           & RMSE & \textbf{0.92} & \textbf{1.83} & \textbf{11.74} & \textbf{0.75} & \textbf{1.52} & \textbf{10.82} & \textbf{0.65} & \textbf{1.34} & \textbf{10.02} & \textbf{0.60} & \textbf{1.24} & \textbf{9.67}  \\
\midrule

\multirow{8}{*}{LSTM}
& \multirow{2}{*}{Palo Alto} & MAE  & 0.63 & 3.32 & 8.13  & 0.51 & 3.06 & 7.61  & 0.47 & 2.92 & 7.34  & 0.44 & 2.79 & 7.08  \\
&                           & RMSE & 0.79 & 4.21 & 10.00 & 0.65 & 3.86 & 9.37  & 0.59 & 3.65 & 8.99  & 0.55 & 3.44 & 8.61  \\
& \multirow{2}{*}{Boulder}  & MAE  & 0.37 & 1.07 & \textbf{2.91}  & 0.35 & 0.99 & \textbf{2.64}  & 0.34 & 0.95 & \textbf{2.46}  & 0.31 & 0.90 & \textbf{2.39}  \\
&                           & RMSE & 0.69 & 1.62 & \textbf{3.95}  & 0.62 & 1.46 & \textbf{3.54}  & 0.56 & 1.33 & \textbf{3.15}  & 0.50 & 1.25 & \textbf{3.07}  \\
& \multirow{2}{*}{Dundee}   & MAE  & 1.05 & 2.02 & 40.20 & 0.97 & 1.93 & 39.90 & 0.93 & 1.89 & 39.66 & 0.91 & 1.85 & 39.66 \\
&                           & RMSE & 1.32 & 2.44 & 43.48 & 1.20 & 2.31 & 43.02 & 1.15 & 2.24 & 42.76 & 1.10 & 2.19 & 42.56 \\
& \multirow{2}{*}{Perth}    & MAE  & 0.61 & 1.39 & 10.25 & 0.54 & 1.24 & 9.49  & 0.50 & 1.14 & 8.71  & 0.48 & 1.08 & 8.60  \\
&                           & RMSE & 0.97 & 1.97 & 13.25 & 0.81 & 1.67 & 12.46 & 0.71 & 1.50 & 11.79 & 0.66 & 1.41 & 11.52 \\
\midrule

\multirow{8}{*}{Transformer}
& \multirow{2}{*}{Palo Alto} & MAE  & \textbf{0.56} & \textbf{2.72} & \textbf{6.02}  & \textbf{0.44} & \textbf{2.45} & \textbf{5.46}  & \textbf{0.41} & \textbf{2.29} & \textbf{5.17}  & \textbf{0.39} & \textbf{2.17} & \textbf{4.99}  \\
&                           & RMSE & 0.76 & \textbf{3.68} & \textbf{8.07}  & 0.62 & \textbf{3.27} & \textbf{7.27}  & 0.56 & \textbf{3.02} & \textbf{6.79}  & 0.51 & \textbf{2.78} & \textbf{6.33}  \\
& \multirow{2}{*}{Boulder}  & MAE  & 0.33 & 0.99 & 3.06  & 0.31 & 0.91 & 2.78  & 0.30 & 0.86 & 2.60  & 0.27 & 0.81 & 2.51  \\
&                           & RMSE & 0.66 & 1.60 & 4.52  & 0.58 & 1.40 & 3.93  & 0.52 & 1.26 & 3.57  & \textbf{0.46} & 1.18 & 3.43  \\
& \multirow{2}{*}{Dundee}   & MAE  & 1.07 & 2.12 & 41.31 & 0.99 & 2.05 & 41.02 & 0.96 & 2.02 & 40.99 & 0.93 & 2.00 & 40.81 \\
&                           & RMSE & 1.34 & 2.52 & 44.27 & 1.23 & 2.41 & 43.73 & 1.17 & 2.36 & 43.53 & 1.13 & 2.31 & 43.12 \\
& \multirow{2}{*}{Perth}    & MAE  & 0.71 & 1.64 & 13.72 & 0.64 & 1.47 & 12.89 & 0.60 & 1.36 & 12.16 & 0.57 & 1.32 & 11.94 \\
&                           & RMSE & 0.99 & 2.07 & 16.31 & 0.84 & 1.79 & 15.64 & 0.75 & 1.63 & 15.02 & 0.71 & 1.55 & 14.85 \\
\bottomrule
\end{tabular}
\end{table}

\begin{table}[htbp]
\centering
\caption{Forecasting results for up to 5-day horizon (1-day steps). Datasets: Palo Alto, Boulder, Dundee, Perth. Lower is better.}
\label{tab:longterm}
\scriptsize
\setlength{\tabcolsep}{2.0pt}
\renewcommand{\arraystretch}{1.05}

\begin{tabular}{l l l c c c c c c c c c c c c c c c c}
\toprule
\textbf{Model} & \textbf{Dataset} & \textbf{Metric} &
\multicolumn{3}{c}{\textbf{1 day}} &
\multicolumn{3}{c}{\textbf{2 days}} &
\multicolumn{3}{c}{\textbf{3 days}} &
\multicolumn{3}{c}{\textbf{4 days}} &
\multicolumn{3}{c}{\textbf{5 days}} \\
\cmidrule(lr){4-6}\cmidrule(lr){7-9}\cmidrule(lr){10-12}\cmidrule(lr){13-15}\cmidrule(lr){16-18}
& &  &
\textbf{St.} & \textbf{Reg.} & \textbf{City} &
\textbf{St.} & \textbf{Reg.} & \textbf{City} &
\textbf{St.} & \textbf{Reg.} & \textbf{City} &
\textbf{St.} & \textbf{Reg.} & \textbf{City} &
\textbf{St.} & \textbf{Reg.} & \textbf{City} \\
\midrule

\multirow{8}{*}{ARIMA}
& \multirow{2}{*}{Palo Alto} & MAE  & 0.88 & 5.15 & 13.43 & 0.79 & 4.78 & 12.74 & 0.77 & 4.54 & 12.14 & 0.75 & 4.32 & 11.69 & 0.75 & 4.21 & 11.43 \\
&                           & RMSE & 1.00 & 6.43 & 17.33 & 0.91 & 6.06 & 16.74 & 0.87 & 5.86 & 16.34 & 0.85 & 5.70 & 16.07 & 0.84 & 5.61 & 15.87 \\
& \multirow{2}{*}{Boulder}  & MAE  & 0.84 & 3.17 & 13.46 & 0.78 & 3.05 & 13.39 & 0.76 & 3.02 & 13.46 & 0.75 & 2.98 & 13.38 & 0.74 & 2.97 & 13.42 \\
&                           & RMSE & 0.99 & 3.42 & 14.05 & 0.89 & 3.27 & 13.94 & 0.85 & 3.19 & 13.89 & 0.83 & 3.15 & 13.83 & 0.81 & 3.14 & 13.85 \\
& \multirow{2}{*}{Dundee}   & MAE  & 1.06 & 2.13 & 33.22 & 1.03 & 2.08 & 32.45 & 1.02 & 2.07 & 32.25 & 1.00 & 2.04 & 31.47 & 1.00 & 2.04 & 30.93 \\
&                           & RMSE & 1.25 & 2.48 & 41.59 & 1.20 & 2.42 & 40.82 & 1.17 & 2.37 & 40.29 & 1.14 & 2.32 & 39.71 & 1.13 & 2.32 & 39.82 \\
& \multirow{2}{*}{Perth}    & MAE  & 0.98 & 2.14 & 20.09 & 0.91 & 1.99 & 18.96 & 0.88 & 1.91 & 18.35 & 0.86 & 1.85 & 17.53 & 0.84 & 1.80 & 16.62 \\
&                           & RMSE & 1.27 & 2.63 & \textbf{22.00} & 1.13 & 2.39 & \textbf{20.90} & 1.07 & 2.28 & \textbf{20.20} & 1.03 & 2.19 & \textbf{19.40} & 0.99 & 2.12 & \textbf{18.60} \\
\midrule

\multirow{8}{*}{XGBoost}
& \multirow{2}{*}{Palo Alto} & MAE  & 0.75 & 6.10 & 17.29 & 0.76 & 5.94 & 17.17 & 0.74 & 5.86 & 17.11 & 0.72 & 5.83 & 17.12 & 0.72 & 5.81 & 17.08 \\
&                           & RMSE & 0.87 & 7.61 & 21.51 & 0.89 & 7.43 & 21.25 & 0.86 & 7.32 & 21.07 & 0.83 & 7.28 & 21.01 & 0.82 & 7.24 & 20.92 \\
& \multirow{2}{*}{Boulder}  & MAE  & 0.70 & 2.28 & 9.42  & 0.60 & 2.12 & 9.35  & 0.54 & 2.05 & 9.39  & 0.53 & 2.02 & 9.43  & 0.51 & 2.00 & 9.36 \\
&                           & RMSE & 0.81 & 2.52 & 10.15 & 0.67 & 2.31 & 9.87  & 0.60 & 2.21 & 9.74  & 0.59 & 2.16 & 9.71  & 0.56 & 2.13 & 9.63 \\
& \multirow{2}{*}{Dundee}   & MAE  & 0.88 & 2.17 & 36.83 & 0.85 & 2.12 & 36.78 & 0.83 & 2.09 & 35.56 & 0.82 & 2.06 & 35.15 & 0.82 & 2.07 & 35.79 \\
&                           & RMSE & 1.07 & 2.42 & 40.85 & 1.01 & 2.35 & 40.19 & 0.96 & 2.26 & 39.50 & 0.94 & 2.23 & 38.46 & 0.94 & 2.22 & 39.06 \\
& \multirow{2}{*}{Perth}    & MAE  & 0.91 & 2.02 & 19.36 & 0.78 & 1.76 & 18.02 & 0.72 & 1.63 & 17.25 & 0.68 & 1.56 & 16.79 & 0.62 & 1.42 & 15.53 \\
&                           & RMSE & 1.10 & \textbf{2.38} & 23.58 & 0.93 & \textbf{2.10} & 22.33 & 0.85 & \textbf{1.97} & 21.43 & 0.81 & \textbf{1.90} & 21.05 & 0.75 & \textbf{1.76} & 19.51 \\
\midrule

\multirow{8}{*}{GRU}
& \multirow{2}{*}{Palo Alto} & MAE  & 0.57 & 4.93 & 13.56 & 0.55 & 4.71 & 13.26 & 0.52 & 4.54 & 13.05 & 0.50 & 4.50 & 12.94 & 0.49 & 4.44 & 12.93 \\
&                           & RMSE & 0.67 & 6.15 & 17.05 & 0.65 & 5.92 & 16.71 & 0.61 & 5.79 & 16.53 & 0.59 & 5.73 & 16.44 & 0.58 & 5.68 & 16.36 \\
& \multirow{2}{*}{Boulder}  & MAE  & 0.51 & 1.31 & 4.05  & 0.39 & 1.08 & 3.65  & 0.37 & 0.99 & 3.48  & 0.32 & 0.93 & 3.41  & 0.30 & 0.90 & 3.23 \\
&                           & RMSE & \textbf{0.66} & 1.60 & 4.95  & \textbf{0.47} & 1.32 & 4.39  & \textbf{0.43} & 1.18 & 4.10  & \textbf{0.38} & 1.11 & 3.99  & \textbf{0.35} & 1.05 & 3.79 \\
& \multirow{2}{*}{Dundee}   & MAE  & 0.74 & 1.76 & \textbf{23.94} & 0.71 & 1.70 & \textbf{23.05} & 0.68 & 1.65 & \textbf{21.92} & 0.65 & 1.62 & \textbf{21.54} & \textbf{0.64} & 1.60 & \textbf{21.48} \\
&                           & RMSE & 0.94 & 2.09 & \textbf{29.59} & 0.89 & 2.01 & \textbf{28.92} & 0.83 & 1.91 & \textbf{27.91} & 0.81 & 1.87 & \textbf{27.10} & 0.80 & 1.86 & \textbf{27.56} \\
& \multirow{2}{*}{Perth}    & MAE  & \textbf{0.89} & \textbf{2.01} & \textbf{18.20} & \textbf{0.76} & \textbf{1.74} & \textbf{16.82} & \textbf{0.70} & \textbf{1.62} & \textbf{16.01} & \textbf{0.67} & \textbf{1.55} & \textbf{15.16} & \textbf{0.60} & \textbf{1.41} & \textbf{14.12} \\
&                           & RMSE & \textbf{1.09} & \textbf{2.38} & 23.28 & \textbf{0.92} & \textbf{2.10} & 22.01 & \textbf{0.84} & \textbf{1.97} & 21.06 & \textbf{0.80} & 1.91 & 20.62 & \textbf{0.74} & \textbf{1.76} & 19.00 \\
\midrule

\multirow{8}{*}{LSTM}
& \multirow{2}{*}{Palo Alto} & MAE  & \textbf{0.55} & \textbf{4.44} & 11.93 & \textbf{0.52} & \textbf{4.21} & 11.61 & \textbf{0.49} & \textbf{4.02} & 11.39 & \textbf{0.47} & \textbf{3.97} & 11.31 & \textbf{0.46} & \textbf{3.91} & 11.27 \\
&                           & RMSE & \textbf{0.65} & \textbf{5.68} & \textbf{15.62} & \textbf{0.62} & \textbf{5.43} & \textbf{15.27} & \textbf{0.58} & \textbf{5.28} & \textbf{15.08} & \textbf{0.56} & \textbf{5.22} & \textbf{14.99} & \textbf{0.55} & \textbf{5.17} & \textbf{14.90} \\
& \multirow{2}{*}{Boulder}  & MAE  & 0.58 & 1.57 & 5.79  & 0.46 & 1.40 & 5.58  & 0.43 & 1.33 & 5.52  & 0.39 & 1.28 & 5.52  & 0.36 & 1.24 & 5.49 \\
&                           & RMSE & 0.71 & 1.84 & 6.63  & 0.53 & 1.60 & 6.28  & 0.49 & 1.49 & 6.10  & 0.44 & 1.44 & 6.05  & 0.42 & 1.38 & 5.93 \\
& \multirow{2}{*}{Dundee}   & MAE  & \textbf{0.72} & \textbf{1.63} & 28.69 & \textbf{0.69} & \textbf{1.57} & 28.32 & \textbf{0.66} & \textbf{1.53} & 27.32 & \textbf{0.64} & \textbf{1.49} & 26.12 & \textbf{0.64} & \textbf{1.48} & 26.86 \\
&                           & RMSE & \textbf{0.88} & \textbf{1.89} & 31.21 & \textbf{0.83} & \textbf{1.80} & 30.57 & \textbf{0.77} & \textbf{1.71} & 29.75 & \textbf{0.75} & \textbf{1.67} & 28.70 & \textbf{0.75} & \textbf{1.67} & 29.30 \\
& \multirow{2}{*}{Perth}    & MAE  & 0.94 & 2.10 & 18.33 & 0.80 & 1.83 & 16.95 & 0.74 & 1.71 & 16.37 & 0.71 & 1.64 & 15.56 & 0.65 & 1.52 & 14.93 \\
&                           & RMSE & 1.12 & 2.50 & 25.06 & 0.96 & 2.24 & 23.86 & 0.89 & 2.11 & 23.03 & 0.85 & 2.05 & 22.62 & 0.79 & 1.92 & 21.15 \\
\midrule

\multirow{8}{*}{Transformer}
& \multirow{2}{*}{Palo Alto} & MAE  & 0.57 & 4.89 & \textbf{11.59} & 0.55 & 4.68 & \textbf{11.11} & 0.52 & 4.55 & \textbf{10.83} & 0.51 & 4.50 & \textbf{10.70} & 0.50 & 4.46 & \textbf{10.62} \\
&                           & RMSE & 0.70 & 6.47 & 16.36 & 0.68 & 6.24 & 16.01 & 0.65 & 6.11 & 15.81 & 0.63 & 6.05 & 15.71 & 0.62 & 6.00 & 15.61 \\
& \multirow{2}{*}{Boulder}  & MAE  & \textbf{0.45} & \textbf{1.14} & \textbf{3.47}  & \textbf{0.35} & \textbf{0.98} & \textbf{3.14}  & \textbf{0.35} & \textbf{0.88} & \textbf{2.89}  & \textbf{0.31} & \textbf{0.83} & \textbf{2.81}  & \textbf{0.28} & \textbf{0.77} & \textbf{2.63}  \\
&                           & RMSE & 0.67 & \textbf{1.54} & \textbf{4.49}  & \textbf{0.47} & \textbf{1.27} & \textbf{4.00}  & 0.44 & \textbf{1.13} & \textbf{3.62}  & 0.39 & \textbf{1.06} & \textbf{3.54}  & 0.36 & \textbf{0.98} & \textbf{3.25}  \\
& \multirow{2}{*}{Dundee}   & MAE  & 1.84 & 4.07 & 93.57 & 1.82 & 4.03 & 93.34 & 1.82 & 4.05 & 94.51 & 1.79 & 3.99 & 93.05 & 1.79 & 3.98 & 93.05 \\
&                           & RMSE & 2.17 & 4.65 & 107.40 & 2.13 & 4.60 & 107.13 & 2.10 & 4.57 & 107.66 & 2.09 & 4.53 & 106.56 & 2.07 & 4.52 & 106.48 \\
& \multirow{2}{*}{Perth}    & MAE  & 0.96 & 2.12 & 19.21 & 0.83 & 1.85 & 17.81 & 0.77 & 1.73 & 17.27 & 0.73 & 1.67 & 16.64 & 0.67 & 1.53 & 15.76 \\
&                           & RMSE & 1.13 & 2.49 & 25.21 & 0.97 & 2.23 & 24.01 & 0.90 & 2.11 & 23.22 & 0.86 & 2.05 & 22.83 & 0.80 & 1.91 & 21.36 \\
\bottomrule
\end{tabular}
\end{table}

For short-term forecasting, performance exhibits clear dataset-dependent behavior. In the Perth dataset, LSTM consistently achieves the lowest errors across all spatial aggregation levels and horizons, dominating station-, regional-, and city-level forecasts. For example, at the station level, LSTM attains MAE values between 0.43 and 0.46 and RMSE values below 1.1 across all short-term horizons, substantially outperforming competing models. In the other datasets, Transformer models demonstrate the strongest overall performance. In Boulder, Transformers achieve the lowest MAE across all spatial levels and horizons, with station-level MAE values around 0.21, compared to 0.26 for ARIMA. Similarly, in Dundee, Transformers dominate regional- and city-level forecasts, while ARIMA attains the lowest station-level MAE but is outperformed by Transformers in terms of RMSE. In Palo Alto, Transformers consistently yield the lowest errors at regional and city scales, whereas XGBoost achieves the best station-level performance, with MAE and RMSE values of approximately 0.87 and 1.01, respectively. Overall, Transformer models emerge as the most reliable short-term forecasting approach across datasets, particularly at regional and city aggregation levels. LSTM exhibits strong dataset-specific performance in Perth, while XGBoost demonstrates competitive behavior only in localized station-level forecasting. ARIMA and GRU generally deliver intermediate performance and do not consistently outperform deep learning models in short-term scenarios.

Moving to the mid-term horizon (up to 8 hours in 2-hour steps; see Table~\ref{tab:midterm}), a clearer separation between model families emerges, although performance remains strongly dataset-dependent. GRU achieves the lowest errors consistently in the Dundee and Perth datasets across all mid-term horizons and spatial aggregation levels, indicating strong robustness in these two cities. In Palo Alto, Transformer models dominate the aggregated forecasts, achieving the best performance at both regional and city scales across 2–8-hour horizons (e.g., city-level MAE decreases from 6.02 at 2 hours to 4.99 at 8 hours). In Boulder, ARIMA performs best at the station and regional levels for most mid-term horizons, whereas LSTM achieves the lowest errors at the city level across the same horizons (e.g., city MAE decreases from 2.91 at 2 hours to 2.39 at 8 hours). XGBoost remains competitive only in selected localized settings, most notably at the Palo Alto station level, where it attains the lowest RMSE, while underperforming at higher aggregation levels and in the other datasets. Across datasets, errors generally decrease as the forecasting horizon increases. Although this may appear counter-intuitive, the effect is well documented in load forecasting under temporal aggregation: short-term EV charging demand is highly volatile, whereas longer horizons smooth out noise and reveal more regular patterns that models can capture more reliably~\cite{b8}.

For long-term forecasting (up to 5 days in 1-day steps; see Table~\ref{tab:longterm}), performance is strongly influenced by both dataset characteristics and spatial aggregation level. In Boulder, Transformer models achieve the lowest MAE across all spatial levels and horizons (e.g., city MAE decreases from 3.47 at 1 day to 2.63 at 5 days), indicating strong effectiveness for this dataset. In Palo Alto, LSTM provides the best performance at the station and regional levels across 1–5 days, while at the city level Transformer attains the lowest MAE (e.g., 11.59 at 1 day decreasing to 10.62 at 5 days) and LSTM achieves the lowest RMSE (e.g., 15.62 decreasing to 14.90). In Dundee, a clear divergence is observed: GRU yields the best city-level forecasts (city MAE around 23.94 at 1 day and 21.48 at 5 days), whereas Transformer performs substantially worse, with city-level errors roughly an order of magnitude higher (city MAE of approximately 93 and RMSE of 106 across horizons). Finally, in Perth, GRU consistently achieves the lowest MAE across station, regional, and city levels, while ARIMA attains the lowest city-level RMSE across all horizons (e.g., 22.00 at 1 day decreasing to 18.60 at 5 days). As in the mid-term case, this reduction in error with increasing horizon can again be attributed to temporal aggregation, which suppresses short-term variability and emphasizes lower-frequency patterns that are more predictable over multi-day horizons.

Overall, the results demonstrate that forecasting performance depends strongly on both the prediction horizon and the spatial aggregation level, with consistent trends observed across datasets alongside meaningful dataset-specific deviations. In short-term forecasting, deep learning models generally outperform classical approaches, with Transformer models performing particularly well at regional and city aggregation levels, while LSTM exhibits strong dataset-specific dominance in Perth. As the forecasting horizon increases, recurrent neural networks emerge as the most reliable models: GRU and LSTM consistently achieve the lowest errors in mid- and long-term forecasting across most datasets and aggregation levels. Transformer models perform competitively in several settings, particularly under coarser spatial aggregation, but also exhibit pronounced sensitivity to dataset characteristics, including notable performance degradation in the Dundee dataset for long-term forecasting. ARIMA remains competitive in isolated cases but does not scale effectively to longer horizons or higher levels of spatial aggregation, while XGBoost shows limited competitiveness outside specific localized scenarios. Taken together, these results indicate that models explicitly designed for sequential learning provide the most robust and reliable performance for EV charging load forecasting across diverse temporal and spatial contexts.

\subsection{A Note on the Transformer Performance in Dundee}
The experimental results show that Transformers perform well in short-term regional and city-level forecasting in Palo Alto and Boulder (10--30 minute horizons), but deteriorate sharply for long-term city-level forecasts in Dundee, where their errors are substantially higher than those of GRU and LSTM (see\ Table~\ref{tab:longterm}).

A plausible explanation is the interaction between dataset size and spatial structure. Dundee has by far the largest number of stations (67) and regions (34) but comparatively few session records (52{,}752 versus 259{,}415 for Palo Alto and 66{,}664 for Perth; see Table~\ref{tab:datasets}). After aggregation to hourly and especially daily series, this yields much shorter and more zero-inflated histories per station and region than in the other cities. In our multi-station setup, the high-capacity Transformer encoder therefore sees relatively few informative examples per spatial unit, making it more prone to overfitting and error accumulation under recursive multi-step forecasting. By contrast, GRU and LSTM impose a stronger inductive bias toward local temporal continuity and appear more robust on this sparse, short-span dataset. Overall, the Dundee results suggest that Transformers are best used when both the time span and the effective sample size per spatial unit are sufficiently large; otherwise, recurrent models may be preferable for long-term forecasts.

\section{Related Work}
\label{sec:related-work}

This section reviews related work on EV charging load forecasting.  We first summarize the broader landscape of forecasting models that have been applied in this domain, highlighting the range of methodological approaches and typical forecasting horizons used in prior studies. We then discuss studies that conduct direct model comparisons and are therefore more closely related to our work.

Traditional models such as ARIMA and SARIMA perform adequately under stable, linear assumptions but struggle with the irregularity of EV charging behavior. For instance, Louie~\cite{b17} applied SARIMA to Washington and San Diego station data, achieving reasonable seasonal modeling but with declining accuracy at longer horizons. 
Machine learning approaches, particularly Random Forests and gradient boosting, often outperform statistical methods in short-term forecasting. Lu et al.~\cite{b18} showed that Random Forests effectively captured Shenzhen station consumption patterns, while CatBoost has demonstrated competitive results for daily EV load forecasting, albeit with challenges in interpretability and hyperparameter tuning.
Deep learning methods, notably LSTM and GRU, have shown clear advantages. Zhu et al.~\cite{b19} reported a 30\% reduction in error using LSTM compared to ANN on Shenzhen data. Unterluggauer et al.~\cite{b20} highlighted the effectiveness of LSTMs for multi-step forecasting in Finland, while Ma and Faye~\cite{b21} achieved near-perfect short-term accuracy with LSTMs.
More recent studies integrate spatial features into the forecasting process. Adinkrah et al.~\cite{b22} reviewed AI-based methods combining clustering with time series forecasting, while Hüttel et al.~\cite{b23} employed Temporal Graph Convolutional Networks (TGCN) to exploit spatial adjacency. Zhang et al.~\cite{b24} further advanced this line of work by applying Graph Neural Networks (GNNs) for hourly station-level forecasting, surpassing traditional baselines. 
Transformers have also gained traction in this domain. Koohfar et al.~\cite{b6} applied Transformer architectures to five years of Boulder station data across horizons from 7 to 90 days, consistently outperforming ARIMA, RNN, and LSTM models.

Beyond methodological proposals, a smaller but growing body of research has focused on experimental comparisons of forecasting models for EV charging. Unlike studies that introduce a single new algorithm, these works benchmark multiple methods under common settings, making them directly comparable to our approach. Reviewing these efforts highlights both their contributions and the distinctive aspects of our study.

\textbf{Hybrid Prophet--LSTM models.}
Wei et al.~\cite{b25} investigate load forecasting for a charging station in southern China using a hybrid model that combines Prophet with LSTM, optimized through a genetic algorithm (GA). Their evaluation, based on daily station-level demand, compares their approach against ARIMA, Prophet, and LSTM baselines. Results show that GA-Prophet-LSTM consistently outperforms the individual models in terms of RMSE and MAPE. While this work provides a clear comparative evaluation, it is limited to a single dataset and daily granularity. In contrast, our study broadens the scope by considering multiple cities, finer temporal resolutions, and spatial aggregation levels.

\textbf{Global versus local forecasting models.}
Van Etten et al.~\cite{b26} propose a large-scale framework for EV charging demand forecasting using \emph{global} models trained across multiple stations. Their study evaluates ARIMA, Transformers, and the N-HiTS architecture on four datasets (London, Palo Alto, Perth, Boulder), using daily horizons of 1, 7, and 30 days. A key novelty lies in testing generalization, where models trained on one set of stations are applied to unseen stations in other regions. Compared to our work, their analysis emphasizes transferability across locations, while ours emphasizes scalability across horizons and spatial aggregation levels (station, region, city). Moreover, their evaluation is restricted to daily data, whereas we include 10-minute, hourly, and daily resolutions. Together, these studies are complementary: their study highlights the potential of global training for cross-station generalization, while our study provides a benchmark across temporal and spatial scales.

\textbf{Real-time availability forecasting.}
Manai et al.~\cite{b27} address the related problem of real-time station availability prediction. Using large-scale datasets from Paris (Belib API) and Estonia (Enefit VOLT), they evaluate models including KNN, logistic regression, Random Forest, SVM, and an Artificial Neural Network (ANN). They also propose an ensemble combining ANN and RF, which outperforms individual baselines across precision, recall, and F1-score. While this study shares our comparative orientation and use of diverse real-world datasets, its target variable differs: station availability (\emph{classification}) rather than energy consumption (\emph{regression}). Moreover, their emphasis is on station-level, user-facing predictions, while our study focuses on multi-horizon load forecasting for infrastructure and grid planning.

\textbf{Forecasting with contextual features.}
Mystakidis et al.~\cite{b28} present a forecasting framework that integrates traffic flow, weather conditions, and user charging behavior to predict hourly EV demand at a residential complex in Trondheim, Norway. To address zero-inflated demand patterns, they introduce a Logarithmic Zero-Inflated (LogZI) regression method, achieving substantial improvements over conventional models in $R^2$, MAE, and RMSE. While their work demonstrates the benefits of incorporating exogenous information and specialized regression techniques, its scope is narrower: one-step-ahead hourly forecasts at a single site. In contrast, our study deliberately restricts to historical load and calendar features, enabling a controlled comparison of model families across multiple horizons and aggregation levels.

To illustrate the heterogeneity of related work, Table~\ref{tab:ev_forecasting_studies} categorizes the studies reviewed in this section with respect to the models used and prediction settings.

In summary, existing studies typically focus either on proposing a specific forecasting architecture (often hybrid or decomposition-based) and demonstrating its superiority on a single dataset and horizon, or on exploring transferability across locations at a fixed temporal resolution. By contrast, our work does not introduce a new model; instead, it provides a structured multi-scale benchmark of widely adopted forecasting methods across four heterogeneous datasets, three spatial aggregation levels, and multiple short-, mid-, and long-term horizons. As such, the results reported here are intended to complement prior state-of-the-art proposals by offering a common reference point against which future models, including graph-based and hybrid architectures, can be evaluated in a consistent manner.

\begin{table}[htbp]
\centering
\caption{Summary of Key Studies on EV Charging Consumption Forecasting}
\label{tab:ev_forecasting_studies}
\renewcommand{\arraystretch}{1.8}
\setlength{\tabcolsep}{8pt}

\adjustbox{max width=\textwidth}{
\begin{tabular}{@{}
>{\raggedright\arraybackslash}p{0.14\textwidth}
>{\raggedright\arraybackslash}p{0.18\textwidth}
>{\raggedright\arraybackslash}p{0.20\textwidth}
>{\raggedright\arraybackslash}p{0.11\textwidth}
>{\raggedright\arraybackslash}p{0.11\textwidth}
>{\raggedright\arraybackslash}p{0.11\textwidth}
@{}}
\hline
\textbf{Study} & \textbf{Model(s) evaluated} & \textbf{Dataset} &
\multicolumn{3}{c}{\textbf{Forecasting Horizon}} \\
\hline
& & &
{\small\textbf{Short-term (minutes)}} &
{\small\textbf{Mid-term (hours)}} &
{\small\textbf{Long-term (days)}} \\
\hline
Our study &
ARIMA, XGBoost, GRU, LSTM, Transformer &
Palo Alto (USA), Boulder (USA), Dundee (UK), Perth (Australia) &
10, 20, 30 &
2, 4, 6, 8 &
1, 2, 3, 4, 5 \\
Louie~\cite{b17} & SARIMA & San Diego (USA) & -- & -- & 1 \\
Lu et al.~\cite{b18} & Random Forests & Shenzhen (China) & 15 & -- & 1 \\
Zhu et al.~\cite{b19} & LSTM, GRU, ANN & Shenzhen (China) & 60 & -- & -- \\
Unterluggauer et al.~\cite{b20} & Multivariate LSTM & Austria & 60 & -- & 1 \\
Ma \& Faye~\cite{b21} & LSTM & Dundee (UK) & 10 & 6 & -- \\
H\"uttel et al.~\cite{b23} & TGCN & Palo Alto (USA) & -- & -- & 1, 7, 30 \\
Zhang et al.~\cite{b24} & GNN & Unspecified & 15, 30, 60 & -- & -- \\
Koohfar et al.~\cite{b6} & Transformer, ARIMA, LSTM, GRU & Boulder (USA) & -- & -- & 7, 30, 60, 90 \\
Wei et al.~\cite{b25} & ARIMA, Prophet, LSTM, GA-Prophet-LSTM & China & -- & -- & 1 \\
Van Etten et al.~\cite{b26} & N-HiTS (global), N-HiTS (local), Transformer, ARIMA, Naive & London (UK), Palo Alto (USA), Perth (Scotland), Boulder (USA) & -- & -- & 1, 7, 30 \\
Mystakidis et al.~\cite{b28} & ZI / LogZI regression; XGBR, LGBMR, HGBR, CBR, RFR, SVR, LSTM, GRU, MLP & Trondheim (Norway) & 60 & -- & -- \\
\hline
\end{tabular}
}
\end{table}

\section{Conclusions and Future Work}
\label{sec:conclusion}
\subsection{Summary and Contributions}

This study conducted a systematic experimental comparison of time series forecasting methods for EV charging demand across different temporal and spatial horizons. Forecasts were evaluated at short-term (up to 30 minutes), mid-term (up to 8 hours), and long-term (up to 5 days) horizons, and at three spatial aggregation levels ranging from individual charging stations to regional and city-wide demand. 

The analysis was based on four publicly available real-world datasets with diverse characteristics. Five forecasting approaches were benchmarked: a classical statistical model (ARIMA), a machine learning model (XGBoost), and three deep learning architectures (GRU, LSTM, and Transformer). All models were trained and evaluated within a unified experimental pipeline, and performance was assessed using MAE and RMSE across identical temporal horizons and spatial aggregation levels. 

The results reveal clear differences across model families and forecasting scenarios. Transformer models are particularly effective in short-term forecasting, especially at regional and city aggregation levels, where they frequently achieve the lowest errors across datasets. ARIMA remains competitive only in isolated station-level cases but does not scale reliably as forecasting horizons increase or spatial aggregation becomes coarser. In contrast, GRU and LSTM consistently deliver the most accurate forecasts in mid- and long-term scenarios, particularly at regional and city scales, highlighting their strength in modeling complex sequential dependencies. Transformer models exhibit mixed behavior beyond the short term and do not consistently outperform recurrent architectures in mid- and long-term forecasting, indicating sensitivity to dataset characteristics and model configuration. XGBoost achieves competitive performance in selected localized scenarios but generally underperforms at higher levels of spatial aggregation.

By evaluating multiple model families across diverse temporal horizons and spatial resolutions, this study provides a comprehensive and reproducible benchmark for EV charging load forecasting. The findings highlight important trade-offs between statistical, machine learning, and deep learning approaches, and underscore the importance of selecting forecasting models that are well matched to the temporal and spatial requirements of the target application.

\subsection{Practical Implications}
The empirical findings of this study have several implications for practitioners involved in the planning and operation of EV charging infrastructure. First, the consistent advantage of Transformer models in short-term regional and city-level forecasting suggests that attention-based architectures are well suited for near real-time grid operation and congestion management, especially when ample historical data are available and demand patterns exhibit relatively smooth short-term dynamics. System operators seeking to anticipate rapid load fluctuations (e.g., over the next 10--30 minutes) may therefore benefit from prioritizing Transformer-type models for aggregated forecasts.

Second, the superior performance of GRU and LSTM networks in mid- and long-term horizons indicates that recurrent architectures remain a robust choice for day-ahead and multi-day planning tasks, such as sizing storage resources, scheduling maintenance, or assessing the impact of EV adoption scenarios on feeder loading. In several datasets, these models reduce MAE substantially compared to statistical baselines, especially at coarser spatial aggregation levels. Finally, the limited and dataset-specific competitiveness of ARIMA and XGBoost in our experiments suggests that, while they remain useful as interpretable baselines and for simpler local forecasts, they may not be sufficient on their own for long-horizon planning in complex urban environments.

\subsection{Limitations}
Several limitations of the present study should be acknowledged. First, we deliberately restrict the input space to historical load and calendar-based features. This design facilitates a controlled comparison of model families, but it also means that important exogenous drivers of EV charging demand, such as weather conditions, electricity prices, traffic patterns, or special events, are not explicitly modeled. As a result, the reported performance should be interpreted as a lower bound on what more richly specified models might achieve in operational settings.

Second, we focus on a set of widely used baseline models (ARIMA, XGBoost, GRU, LSTM, Transformer) and do not include more specialized state-of-the-art architectures such as graph neural networks, decomposition-based hybrids, or global forecasting models tailored to EV data ~\cite{b29}~\cite{b30}. These approaches are instead reviewed in Section~\ref{sec:related-work}, and our results should be viewed as a benchmark for comparing such models rather than as an exhaustive state-of-the-art comparison.

 Finally, the use of recursive multi-step forecasting, although operationally realistic, introduces error accumulation, which may affect some architectures (e.g., Transformers on sparse datasets) more than others. Future work can address these limitations by incorporating exogenous variables, exploring hybrid and graph-based models, and performing more extensive robustness analyses.

\subsection{Future Work}

Several directions remain open for future research. First, while fixed configurations were used to ensure comparability, further improvements could be achieved through systematic hyperparameter optimization and exploration of deeper or more specialized architectures.

Second, the study focused exclusively on historical load and calendar-based features. Incorporating exogenous variables such as weather, traffic patterns, electricity prices, or station location characteristics may improve explanatory power and predictive accuracy.

Finally, to better understand generalization and reduce data requirements, future research could explore explainable AI (XAI) and transfer learning techniques. These would, on the one hand, provide insights into the main parameters that affect each model’s prediction quality, and, on the other, allow knowledge gained from forecasting in one city’s charging stations to be applied to another, potentially improving scalability and adaptability across regions.

\section*{Acknowledgment}
This work was supported in part by the Horizon Europe Research and Innovation Programme of the European Union under grant agreement No. 101070416 (Green.Dat.AI; https://greendatai.eu) and in part by the University of Piraeus Research Center.

\end{document}